\documentclass[10pt,twocolumn,letterpaper]{article}

\usepackage{cvpr}
\usepackage{times}
\usepackage{epsfig}
\usepackage{graphicx}
\usepackage{amsmath}
\usepackage{amssymb}
\usepackage{subfigure}
\usepackage{multirow}
\usepackage{textcomp}
\usepackage{authblk}

\usepackage[breaklinks=true,bookmarks=false]{hyperref}

\cvprfinalcopy % *** Uncomment this line for the final submission

\def\cvprPaperID{****} % *** Enter the CVPR Paper ID here

\begin{document}

%%%%%%%%% TITLE
\title{DeepFashion2: A Versatile Benchmark for Detection, Pose Estimation, \\Segmentation
	and Re-Identification of Clothing Images}

\author[1]{Yuying Ge}
\author[1]{Ruimao Zhang}
\author[2]{Lingyun Wu}
\author[1]{Xiaogang Wang}
\author[1]{Xiaoou Tang}
\author[1]{Ping Luo}
\affil[1]{The Chinese University of Hong Kong}
\affil[2]{SenseTime Research}

\maketitle
%\thispagestyle{empty}

%%%%%%%%% ABSTRACT
\begin{abstract}
  Understanding fashion images has been advanced by benchmarks with rich annotations such as DeepFashion, whose labels include clothing categories, landmarks, and consumer-commercial image pairs.
  % becomes a hot research topic in recent years.
  %
  However, DeepFashion has nonnegligible issues such as single clothing-item per image, sparse landmarks (4$\sim$8 only), and no per-pixel masks, making it had significant gap from real-world scenarios.
  % that possess large variations of scale, occlusion, zooming, and human pose.
  %
  We fill in the gap by presenting DeepFashion2 to address these issues.
  %in its prior release.
  It is a versatile benchmark of four tasks including clothes detection, pose estimation, segmentation, and retrieval.
  %
  %Their annotations are strictly examined by human, facilitating model development for practical applications.
  %
  It has 801K clothing items where each item has rich annotations such as style, scale, viewpoint, occlusion, bounding box, dense landmarks (\eg 39 for `long sleeve outwear' and 15 for `vest'), and masks. There are also
  %totally 801K bboxes, 801K landmarks, 801K masks, and
  873K Commercial-Consumer clothes pairs.
  % category has more than XX dense landmark definition.
  The annotations of DeepFashion2 are much larger than its counterparts such as 8$\times$ of FashionAI Global Challenge.
  %in terms of annotated landmarks.
  A strong baseline is proposed, called Match R-CNN, which builds upon Mask R-CNN to solve the above four tasks in an end-to-end manner. Extensive evaluations are conducted with different criterions in DeepFashion2.
  DeepFashion2 Dataset will be released at : \href{https://github.com/switchablenorms/DeepFashion2}{https://github.com/switchablenorms/DeepFashion2} 
\end{abstract}

%%%%%%%%% BODY TEXT
\section{Introduction}

Fashion image analyses are active research topics in recent years because of their huge potential in industry.
With the development of fashion datasets  ~\cite{C:CCP,C:WTB,C:DARN,C:Attributes,C:deepfashion,C:MVCAD,C:ModaNet,C:fashionAI}, significant progresses have been achieved in this area ~\cite{C:SemanticAtt,C:Unconstrained,C:FashionGrammer,C:PaperDoll,C:2018retrieval,C:2017retrieval}.

\begin{figure}
	\centering
	\includegraphics[width=1.0\linewidth]{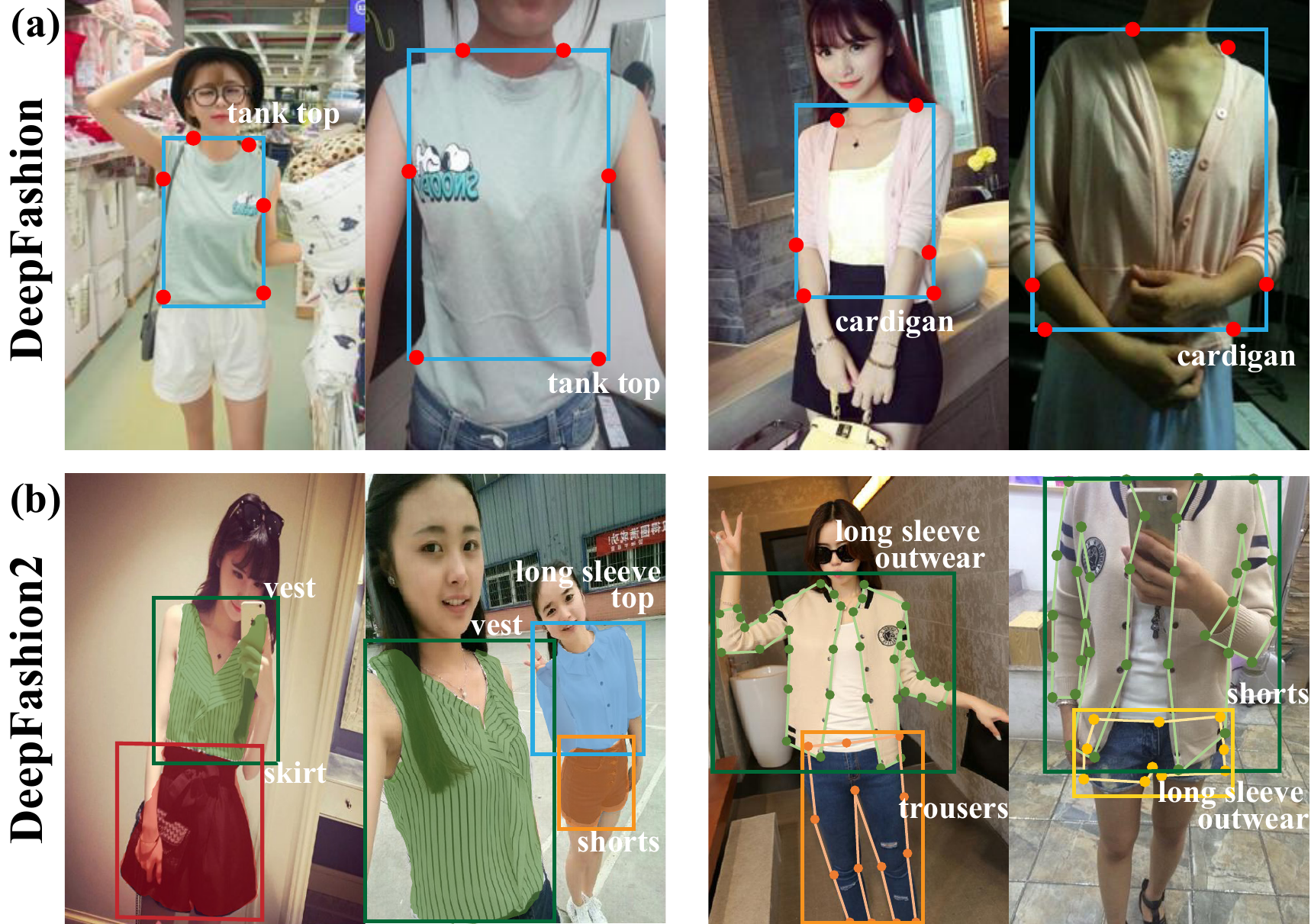}
	\caption{Comparisons between (a) DeepFashion and (b) DeepFashion2. (a) only has single item per image, which is annotated with $4\sim8$ sparse landmarks. {The bounding boxes are estimated from the labeled landmarks, making them noisy.} In (b), each image has minimum single item while maximum 7 items. Each item is manually labeled with bounding box, mask, dense landmarks (20 per item on average), and commercial-customer image pairs.}
	\label{fig:df1_vs_df2}
\end{figure}

However, understanding fashion images remains a challenge in real-world applications, because of large deformations, occlusions, and discrepancies of clothes across domains between consumer and commercial images.
Some challenges can be rooted in the gap between the recent benchmark and the practical scenario.
For example, the existing largest fashion dataset, DeepFashion \cite{C:deepfashion}, has its own drawbacks such as single clothing item per image, sparse landmark and pose definition (every clothing category shares the same definition of $4\sim8$ keypoints), and no per-pixel mask annotation as shown in Fig.\ref{fig:df1_vs_df2}(a).

To address the above drawbacks, this work presents DeepFashion2, a large-scale benchmark with comprehensive tasks and annotations of fashion image understanding.
DeepFashion2 contains 491K images of 13 popular clothing categories. A full spectrum of tasks are defined on them including clothes detection and recognition, landmark and pose estimation, segmentation, as well as verification and retrieval.
All these tasks are supported by rich annotations.
For instance, DeepFashion2 totally has 801K clothing items, where each item in an image is labeled with scale, occlusion, zooming, viewpoint, bounding box, dense landmarks, and per-pixel mask, as shown in Fig.\ref{fig:df1_vs_df2}(b).
%, and pair of images of identical item from consumer and commercial store.
%
These items can be grouped into 43.8K clothing identities, where a clothing identity represents the clothes that have almost the same cutting, pattern, and design. The images of the same identity are taken by both customers and commercial shopping stores. An item from the customer and an item from the commercial store forms a pair. There are 873K pairs that are 3.5 times larger than DeepFashion.
%.
%
The above thorough annotations enable developments of strong algorithms to understand fashion images.
%
%Fig.\ref{fig:df1_vs_df2}(b) shows examples of DeepFashion2.
%

This work has three main \textbf{contributions}.
(1) We build a large-scale fashion benchmark with comprehensive tasks and annotations, to facilitate fashion image analysis.
DeepFashion2 possesses the richest definitions of tasks and the largest number of labels. Its annotations are at least 3.5$\times$ of DeepFashion \cite{C:deepfashion}, 6.7$\times$ of ModaNet \cite{C:ModaNet}, and 8$\times$ of FashionAI \cite{C:fashionAI}.
(2) A full spectrum of tasks is carefully defined on the proposed dataset. For example, to our knowledge, clothing pose estimation is presented for the first time in the literature
by defining landmarks and poses of 13 categories that are more diverse and fruitful than human pose.
(3) With DeepFashion2, we extensively evaluate Mask R-CNN \cite{C:MaskRCNN} that is a recent advanced framework for visual perception. A novel Match R-CNN is also proposed to aggregate all the learned features from clothes categories, poses, and masks to solve clothing image retrieval in an end-to-end manner.
%build upon it
DeepFashion2 and implementations of Match R-CNN will be released.

%-------------------------------------------------------------------------
\subsection{Related Work}

\textbf{Clothes Datasets.} Several clothes datasets have been proposed such as~\cite{C:CCP,C:WTB,C:DARN,C:deepfashion,C:ModaNet,C:fashionAI}
as summarized in Table~\ref{tab:DatasetComparison}. They vary in size as well as amount and type of annotations.
For example, WTBI~\cite{C:WTB} and DARN \cite{C:DARN} have 425K and 182K images respectively. They scraped category labels from metadata of the collected images from online shopping websites, making their labels noisy.
% For example, DARN has 179 categories that are noisy.
% such as product keywords and descriptions,
In contrast, CCP~\cite{C:CCP}, DeepFashion \cite{C:deepfashion}, and ModaNet \cite{C:ModaNet} obtain category labels from human annotators.
Moreover, different kinds of annotations are also provided in these datastes.
For example, DeepFashion labels 4$\sim$8 landmarks (keypoints) per image that are defined on the functional regions of clothes (\eg `collar'). The definitions of these sparse landmarks are shared across all categories, making them difficult to capture rich variations of clothing images. Furthermore, DeepFashion does not have mask annotations.
% is introduced in fashion dataset.
%
By comparison, ModaNet~\cite{C:ModaNet} has street images with masks (polygons) of single person but without landmarks.
Unlike existing datasets, DeepFashion2 contains 491K images and 801K instances of landmarks, masks, and bounding boxes, as well as 873K pairs. It is the most comprehensive benchmark of its kinds to date.

\textbf{Fashion Image Understanding.} There are various tasks that analyze clothing images such as clothes detection~\cite{C:SemanticAtt,C:deepfashion}, landmark prediction~\cite{C:FashionLandmark,C:Unconstrained,C:FashionGrammer}, clothes segmentation~\cite{C:PaperDoll,C:CCP,J:FashionParsing}, and retrieval~\cite{C:DARN,C:WTB,C:deepfashion}.
However, a unify benchmark and framework to account for all these tasks is still desired.
DeepFashion2 and Match R-CNN fill in this blank.
%
%Our work is also related to these articles.
%
We report extensive results for the above tasks with respect to different variations, including scale, occlusion, zoom-in, and viewpoint.
%
%large-scale and detailed results when considering various degrees of clothing size, occlusion, zooming and human pose.
%
For the task of clothes retrieval, unlike previous methods \cite{C:WTB,C:DARN} that performed image-level retrieval,
DeepFashion2 enables instance-level retrieval of clothing items.
% in retrieval.
We also present a new fashion task called clothes pose estimation, which is inspired by human pose estimation to predict clothing landmarks and skeletons for 13 clothes categories.
This task helps improve performance of fashion image analysis in real-world applications.

\begin{table}[t]
	\scriptsize
	\centering{
		\begin{tabular}{p{0.9cm}p{0.7cm}p{0.7cm}p{0.6cm}p{0.7cm}p{0.7cm}p{0.6cm}}
			\hline
			\quad  & \rotatebox{30}{WTBI} & \rotatebox{30}{DARN} & \rotatebox{30}{DeepFashion} & \rotatebox{30}{ModaNet} & \rotatebox{30}{FashionAI} &\rotatebox{30}{DeepFashion2}\\
			\hline
			year    &  2015\cite{C:WTB}  &    2015\cite{C:DARN}    &   2016\cite{C:deepfashion}   &    2018\cite{C:ModaNet}  &   2018\cite{C:fashionAI} & now    \\
			\hline
			$\#$images  & 425K & 182K & 800K& 55K&357K &491K \\
			\hline
			$\#$categories  & 11 & 20 & 50 & 13 &41 &13 \\
			\hline
			$\#$bboxes  & 39K &7K &$\times$ &$\times$ &$\times$ &801K\\
			\hline
			$\#$landmarks  & $\times$ &$\times$ &120K &$\times$ &100K &801K \\
			\hline
			$\#$masks  & $\times$ & $\times$ & $\times$ &119K &$\times$ & 801K \\
			\hline
			$\#$pairs  & 39K &91K &251K & $\times$ &$\times$ & 873K \\
			\hline
	\end{tabular}}
	\vspace{3pt}
	\caption{Comparisons of DeepFashion2 with the other clothes datasets. The rows represent number of images, bounding boxes, landmarks, per-pixel masks, and consumer-to-shop pairs respectively. Bounding boxes inferred from other annotations are not counted.}
	\label{tab:DatasetComparison}
	\vspace{-9pt}
\end{table}

\begin{figure*}
	\centering
	\includegraphics[width=1.0\linewidth]{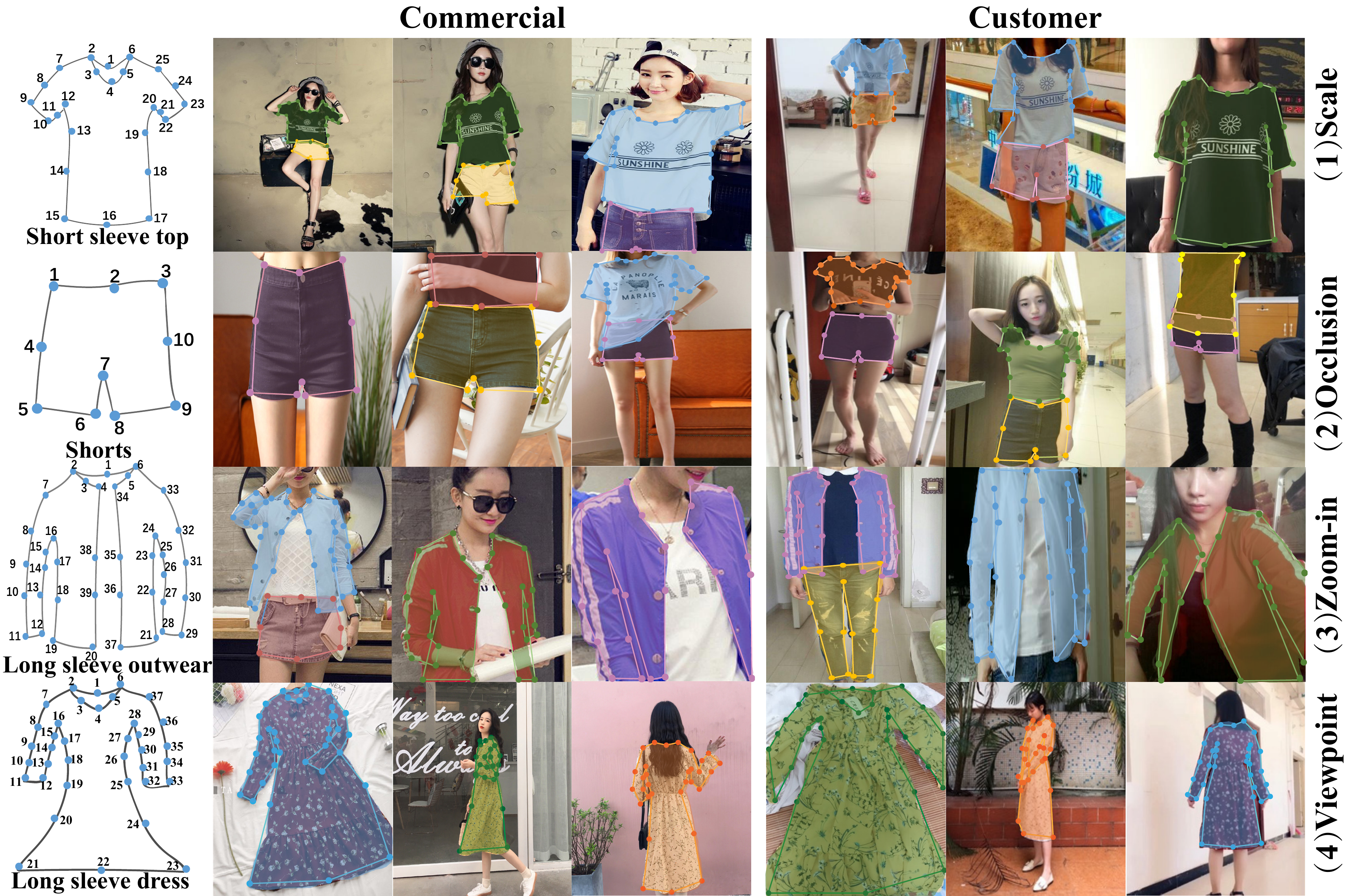}
	\caption{\textbf{Examples of DeepFashion2.} The first column shows definitions of dense landmarks and skeletons of four categories.
		From (1) to (4), each row represents clothes images with different variations including `scale', `occlusion', `zoom-in', and `viewpoint'.
		At each row, we partition the images into two groups, the left three columns represent clothes from commercial stores, while the right three columns are from customers.
		In each group, the three images indicate three levels of difficulty with respect to the corresponding variation, including (1) `small', `moderate', `large' scale, (2) `slight', `medium', `heavy' occlusion, (3) `no', `medium', `large' zoom-in, (4) `not on human', `side', `back' viewpoint.
		Furthermore, at each row, the items in these two groups of images are from the same clothing identity but from two different domains, that is, commercial and customer.
		The items of the same identity may have different styles such as color and printing.
		Each item is annotated with landmarks and masks.
		%shop images and their corresponding user-taken photos with the same clothing identity but differ in styles such as color and pattern and also have different properties. Each clothing item is annotated with mask, dense landmarks and pose.
	}
	\label{fig:overall}
\end{figure*}

\section{DeepFashion2 Dataset and Benchmark}

%\subsection{Overview}
\textbf{Overview.}
DeepFashion2 has four unique characteristics compared to existing fashion datasets.
%Comparison with previous fashion dataset, DeepFashion2 has the following advantages:
%
(1) \textit{Large Sample Size.} It contains 491K images of 43.8K clothing identities of interest (unique garment displayed by shopping stores). On average, each identity has 12.7 items with different styles such as color and printing.
DeepFashion2 contained 801K items in total. It is the largest fashion database to date.
%from 13 carefully defined categories where each image is labeled with multiple clothing items.
%Fig.\ref{} shows several examples.
%
Furthermore, each item is associated with various annotations as introduced above.

(2) \textit{Versatility.} DeepFashion2 is developed for multiple tasks of fashion understanding.
Its rich annotations support clothes detection and classification, dense landmark and pose estimation, instance segmentation, and cross-domain instance-level clothes retrieval.

(3) \textit{Expressivity.} This is mainly reflected in two aspects.
First, multiple items are present in a single image, unlike DeepFashion where each image is labeled with at most one item.
Second, we have 13 different definitions of landmarks and poses (skeletons) for 13 different categories. There is 23 defined landmarks for each category on average. Some definitions are shown in the first column of Fig.\ref{fig:overall}.
These representations are different from human pose and are not presented in previous work.
They facilitate learning of strong clothes features that satisfy real-world requirements.

(4) \textit{Diversity.} We collect data by controlling their variations in terms of four properties including scale, occlusion, zoom-in, and viewpoint as illustrated in Fig.\ref{fig:overall}, making DeepFashion2 a challenging benchmark.
For each property, each clothing item is assigned to one of three levels of difficulty.
Fig.\ref{fig:overall} shows that each identity has high diversity where its items are from different difficulties.

\textbf{Data Collection and Cleaning.}
Raw data of DeepFashion2 are collected from two sources including DeepFashion \cite{C:deepfashion} and online shopping websites.
In particular, images of each consumer-to-shop pair in DeepFashion are included in DeepFashion2, while the other images are removed. We further crawl a large set of images on the Internet from both commercial shopping stores and consumers.
To clean up the crawled set, we first remove shop images with no corresponding consumer-taken photos.
Then human annotators are asked to clean images that contain clothes with large occlusions, small scales, and low resolutions.
Eventually we have 491K images of 801K items and 873K commercial-consumer pairs.
% where each item is at least involved in one pair.

\begin{figure*}
	\centering
	\includegraphics[width=0.9\linewidth]{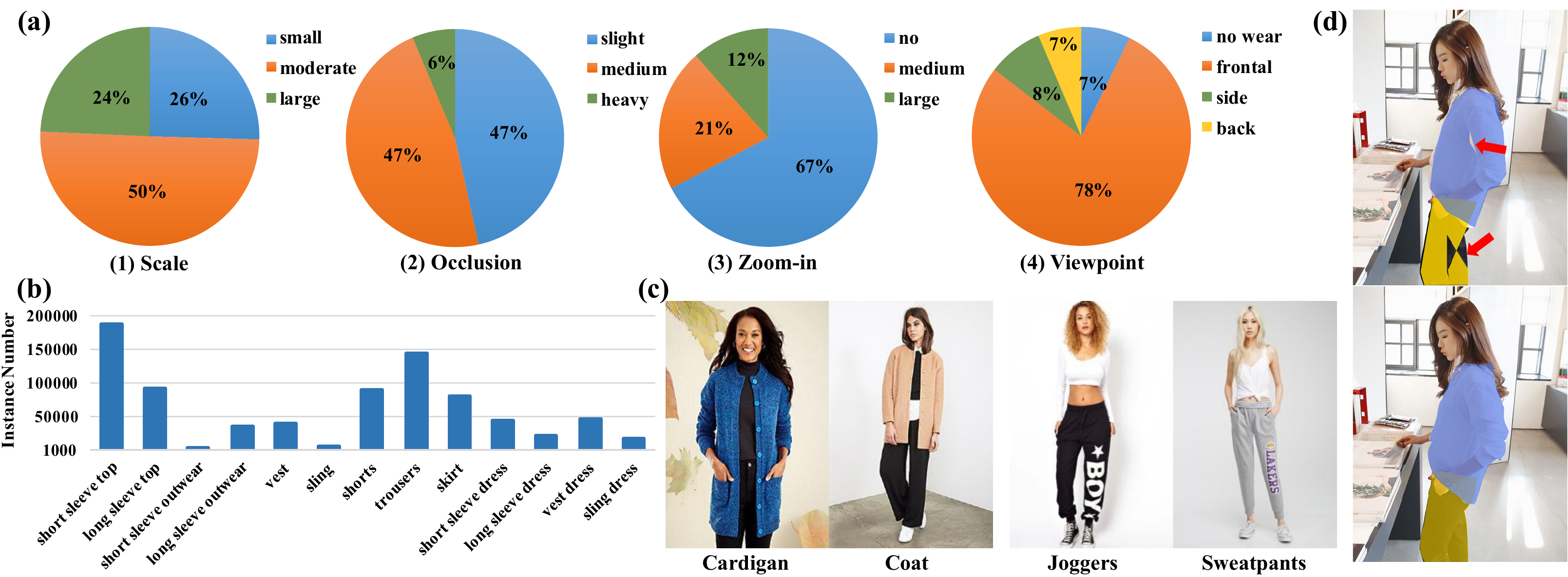}
	\caption{(a) shows the statistics of different variations in DeepFashion2. (b) is the numbers of items of the 13 categories in DeepFashion2. (c) shows that categories in DeepFashion \cite{C:deepfashion} have ambiguity. For example, it is difficult to distinguish between `cardigan' and `coat', and between `joggers' and `sweatpants'. They result in ambiguity when labeling data. (d) \textbf{Top:} masks may be inaccurate when complex poses are presented.
		\textbf{Bottom:}  the masks will be refined by human.  }
	\label{fig:properties}
\end{figure*}

\textbf{Variations.}
We explain the variations in DeepFashion2.
Their statistics are plotted in Fig.\ref{fig:properties}.
(1) \textbf{\textit{Scale.}} We divide all clothing items into three sets, according to the proportion of an item compared to the image size, including `small' ($<10\%$), `moderate' ($10\%\sim40\%$), and `large' ($>40\%$).
Fig.\ref{fig:properties}(a) shows that only  $50\%$ items have moderate scale.
%, a proper size for fashion images.
%Fig.\ref{statistics historgram}(b) shows that more than $70\%$ items have small or medium scale, occupying $<40\%$ regions of their images.£¨large scale is indeed more difficult from experiments)
%
%
(2) \textbf{\textit{Occlusion.}} An item with occlusion means that its region is occluded by hair, human body, accessory or other items.
Note that an item with its region outside the image does not belong to this case.
Each item is categorized by the number of its landmarks that are occluded, including
%
%we define the occlusion of a image as follows:
%
`partial occlusion'($<20\%$ occluded keypoints), `heavy occlusion' ($>50\%$ occluded keypoints), `medium occlusion' (otherwise).
More than 50\% items have medium or heavy occlusions as summarized in Fig.\ref{fig:properties}.
(3) \textbf{\textit{Zoom-in.}} An item with zoom-in means that its region is outside the image.
%, and the camera only captures part of it with close shoot.
%
%According to the proportion of the number of edge landmarks,
This is categorized by the number of landmarks outside image.
We define `no', `large' ($>30\%$), and `medium' zoom-in.
We see that more than 30\% items are zoomed in.
(4) \textbf{\textit{Viewpoint.}} We divide all items into four partitions including 7\% clothes that are not on people, 78\% clothes on people from frontal viewpoint,
15\% clothes on people from side or back viewpoint.

\subsection{Data Labeling}

\textbf{Category and Bounding Box.}
Human annotators are asked to draw a bounding box and assign a category label for each clothing item.
%
%Then a pre-defined category label is assigned to each annotated instance.
%
DeepFashion \cite{C:deepfashion} defines 50 categories but half of them contain less than 5\textperthousand~number of images.
Also, ambiguity exists between 50 categories making data labeling difficult as shown in Fig.\ref{fig:properties}(c).
By grouping categories in DeepFashion, we derive 13 popular categories without ambiguity.
%
%They are defined by grouping categories of similar appearance and removing those rarely presented in practice.
%
The numbers of items of 13 categories are shown in Fig.\ref{fig:properties}(b).

\textbf{Clothes Landmark, Contour, and Skeleton.}
As different categories of clothes (\eg upper- and lower-body garment) have different deformations and appearance changes, we represent each category by defining its pose, which is a set of landmarks as well as contours and skeletons between landmarks. They capture shapes and structures of clothes.
%Fig.\ref{fig:apendix} in Appendix provides pose definitions for 13 categories, which
Pose definitions are not presented in previous work and are significantly different from human pose.
For each clothing item of a category, human annotations are asked to label landmarks following these instructions.

Moreover, each landmark is assigned one of the two modes, `visible' or `occluded'.
We then generate contours and skeletons automatically by connecting landmarks in a certain order.
To facilitate this process, annotators are also asked to distinguish landmarks into two types, that is, ¡®contour point¡¯ or ¡®junction point¡¯.
The former one refers to keypoints at the boundary of an item, while
the latter one is assigned to keypoints in conjunction \eg `endpoint of strap on sling'.
The above process controls the labeling quality, because the generated skeletons help the annotators reexamine whether the landmarks are labeled with good quality.
% is qualified.
%
In particular, only when the contour covers the entire item, the labeled results are eligible,
otherwise keypoints will be refined.

\textbf{Mask.}
We label per-pixel mask for each item in a semi-automatic manner with two stages.
The first stage automatically generates masks from the contours. In the second stage, human annotators are asked to refine the masks, because the generated masks may be not accurate when complex human poses are presented. As shown in Fig.\ref{fig:properties}(d), the mark is inaccurate when an image is taken from side-view of people crossing legs.
%
%shows an example.
The masks will be refined by human.

\textbf{Style.}
As introduced before, we collect 43.8K different clothing identities where each identity has 13 items on average. These items are further labeled with different styles such as color, printing, and logo. %Other clothing items are labelled with style 0 and do not construct pairs.
Fig.\ref{fig:overall} shows that a pair of clothes that have the same identity could have different styles.

\subsection{Benchmarks}

We build four benchmarks by using the images and labels from DeepFashion2.
For each benchmark, there are 391K images for training, 34K images for validation and 67K images for test.

\textbf{Clothes Detection.}
This task detects clothes in an image by predicting bounding boxes and category labels.
%to each detected clothing item.
%
The evaluation metrics are the bounding box's average precision $\mathrm{AP}_{\mathrm{box}}$, $\mathrm{AP^{IoU=0.50}_{box}}$, and $\mathrm{AP^{IoU=0.75}_{box}}$ by following COCO~\cite{C:COCO}.
% are employed in this task. We calculate $AP$, $AP^{IoU=0.50}$, $AP^{IoU=0.75}$ and $IoU$ is computed over boxes.

\textbf{Landmark Estimation.} This task aims to predict landmarks for each detected clothing item in an each image.
%
%The previous defined 13 categories are group to 9 categories, namely top with sleeve, outwear, vest, sling, pants, skirt, dress with sleeve, vest dress, sling dress.
%
Similarly, we employ the evaluation metrics used by COCO for human pose estimation by
% are employed.
%
calculating the average precision for keypoints $\mathrm{AP_{pt}}$, $\mathrm{AP^{OKS=0.50}_{pt}}$, and $\mathrm{AP^{OKS=0.75}_{pt}}$, where $\mathrm{OKS}$ indicates the object landmark similarity.
% which plays the same role as the $IoU$.
%
%The per-landmark standard deviation $\sigma_{i}$ with respect to object scale $s$ for each landmark type is measured using 5000 redundantly annotated images in val. $\sigma_{i}$ varies for different landmarks: for example, landmarks on the hemline have larger $\sigma$s compared with landmarks on collars.

\textbf{Segmentation.}
This task assigns a category label (including background label) to each pixel in an item.
%
%The training set contain 391K images with pixel information \textit{generated by landmarks}.
%
%34K images in the validation set and 67K images are fully annotated by human annotators.
%
The evaluation metrics is the average precision including $\mathrm{AP_{mask}}$, $\mathrm{AP^{IoU=0.50}_{mask}}$, and $\mathrm{AP^{IoU=0.75}_{mask}}$ computed over masks.

\textbf{Commercial-Consumer Clothes Retrieval.}
%
%Different from DeepFashion that evaluates image-level retrieval,
Given a detected item from a consumer-taken photo, this task aims to search the commercial images in the gallery for the items that are corresponding to this detected item.
% from an consumer-taken image.
%
This setting is more realistic than DeepFashion \cite{C:deepfashion}, which assumes ground-truth bounding box is provided.
%which only performs image-level retrieval.
%
In this task, top-k retrieval accuracy is employed as the evaluation metric.
%, which combines both detection and verification results.
%
%
We emphasize the retrieval performance while still consider the influence of detector.
If a clothing item fails to be detected, this query item is counted as missed.
In particular, we have more than 686K commercial-consumer clothes pairs in the training set.
% with at least one matching clothing item pair each image in the training set.
%
In the validation set, there are $10,990$ consumer images with $12,550$ items as a query set, and $21,438$ commercial images with $37,183$ items as a gallery set.
In the test set, there are $21,550$ consumer images with $24,402$ items as queries, while $43,608$ commercial images with $75,347$ items in the gallery.

\section{Match R-CNN}\label{sec:implementation}

We present a strong baseline model built upon Mask R-CNN \cite{C:MaskRCNN} for DeepFashion2, termed Match R-CNN,
%To address the problem of fashion understanding, we propose
%
which is an end-to-end training framework that jointly learns clothes detection, landmark estimation, instance segmentation, and consumer-to-shop retrieval.
The above tasks are solved by using different streams and stacking a Siamese module on top of these streams to aggregate learned features.
% from predicted bounding boxes, poses, and masks for clothes retrieval.
%

\begin{figure}[t]
	\begin{center}
		\includegraphics[width=1.1\linewidth,height=60mm]{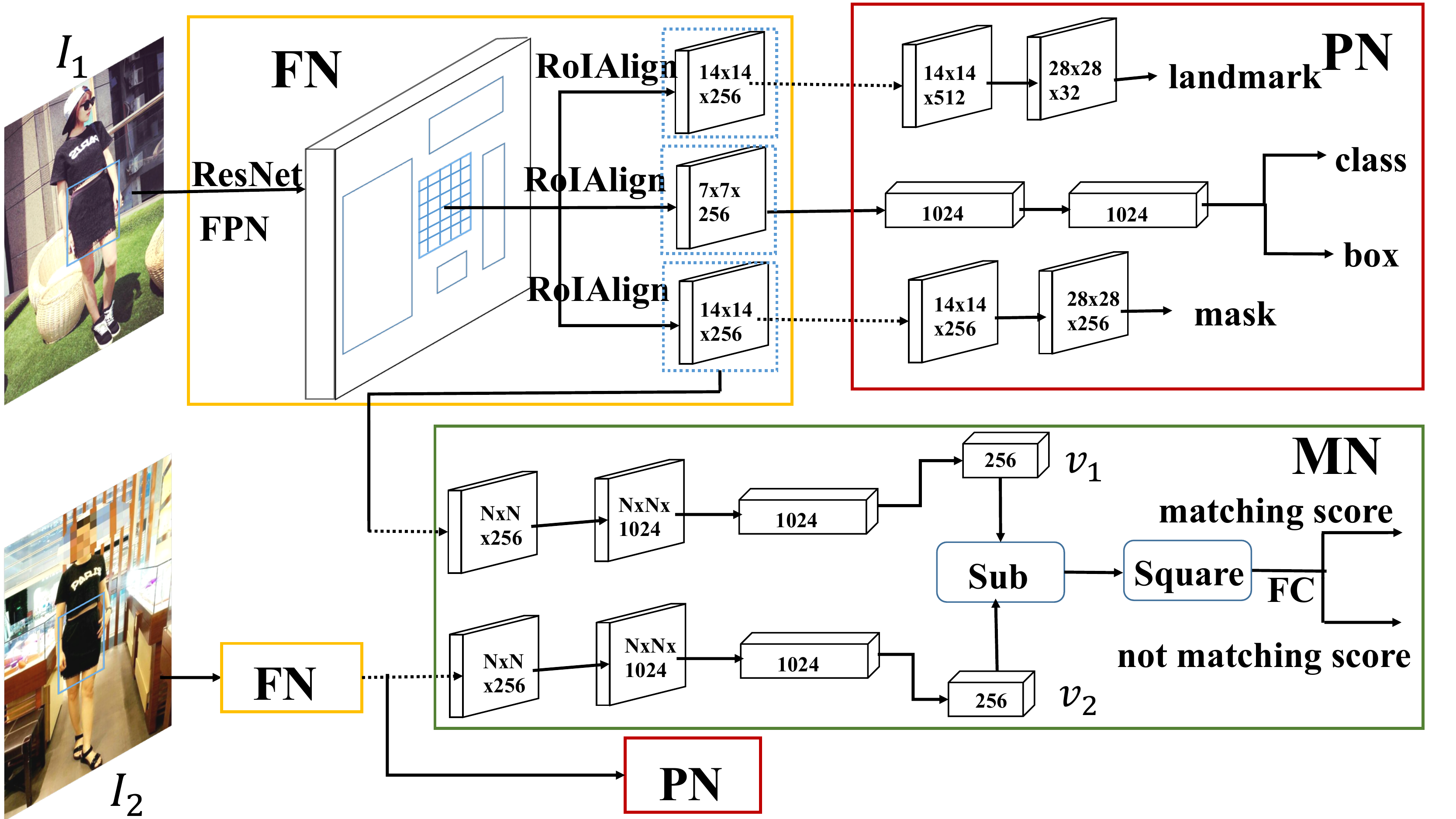}
	\end{center}
	\caption{\textbf{Diagram of Match R-CNN} that contains three main components including a feature extraction network (FN), a perception network (PN), and a match network (MN).}
	\label{fig:Model}
\end{figure}

As shown in Fig.\ref{fig:Model}, Match R-CNN employs two images $I_1$ and $I_2$ as inputs. Each image is passed through three main components including a Feature Network (FN), a Perception Network (PN), and a Matching Network (MN).
In the first stage, FN contains a ResNet-FPN \cite{C:FPN} backbone, a region proposal network (RPN) \cite{C:faster-rcnn} and RoIAlign module.
An image is first fed into ResNet50 to extract features, which are then fed into a FPN that uses a top-down architecture with lateral connections to build a pyramid of feature maps.
%
%After RPN generates RoIs, 
RoIAlign extracts features from different levels of the pyramid map.
% according to their scale.
%
%After the first stage, features from each candidate box are extracted.

%
In the second stage, PN contains three streams of networks including landmark estimation, clothes detection, and mask prediction as shown in Fig.\ref{fig:Model}.
The extracted RoI features after the first stage are fed into three streams in PN separately.
The clothes detection stream has two hidden fully-connected (fc) layers, one fc layer for classification, and one fc layer for bounding box regression. The stream of landmark estimation has 8 `conv' layers and 2 `deconv' layers to predict landmarks. Segmentation stream has 4 `conv' layers, 1 `deconv' layer, and another `conv' layer to predict masks.
%
%Extracted RoI features after the first stage are fed into three streams in PN separately for class and box, landmark, and mask results.

%
\begin{table*}\small
	\scalebox{0.9}{
		\begin{tabular}{l|ccc|ccc|ccc|ccc|c}
			\hline
			&\multicolumn{3}{c|}{scale}&\multicolumn{3}{c|}{occlusion}&\multicolumn{3}{c|}{zoom-in}&\multicolumn{3}{c|}{viewpoint}&\multicolumn{1}{c}{overall}  \\
			%\hline
			\quad &small&moderate&large&slight&medium&heavy&no&medium&large&no wear&frontal&side or back \\
			\hline
			$\mathrm{AP}_{\mathrm{box}}$    &0.604&\textbf{0.700}&0.660&\textbf{0.712}&0.654&0.372&\textbf{0.695}&0.629&0.466&0.624&\textbf{0.681}&0.641&0.667\\
			$\mathrm{AP^{IoU=0.50}_{box}}$ &0.780&\textbf{0.851}&0.768&\textbf{0.844}&0.810&0.531&\textbf{0.848}&0.755&0.563&0.713&\textbf{0.832}&0.796&0.814\\
			$\mathrm{AP^{IoU=0.75}_{box}}$ &0.717&\textbf{0.809}&0.744&\textbf{0.812}&0.768&0.433&\textbf{0.806}&0.718&0.525&0.688&\textbf{0.791}&0.744&0.773\\
			\hline
	\end{tabular}}
	\vspace{3pt}
	\caption{\textbf{Clothes detection} of Mask R-CNN \cite{C:MaskRCNN} on different validation subsets, including scale, occlusion, zoom-in, and viewpoint. The evaluation metrics are $\mathrm{AP}_{\mathrm{box}}$, $\mathrm{AP^{IoU=0.50}_{box}}$, and $\mathrm{AP^{IoU=0.75}_{box}}$. The best performance of each subset is bold.}
	\label{tab:detection}
\end{table*}

\begin{figure*}[t]
	\begin{center}
		\includegraphics[width=1\linewidth]{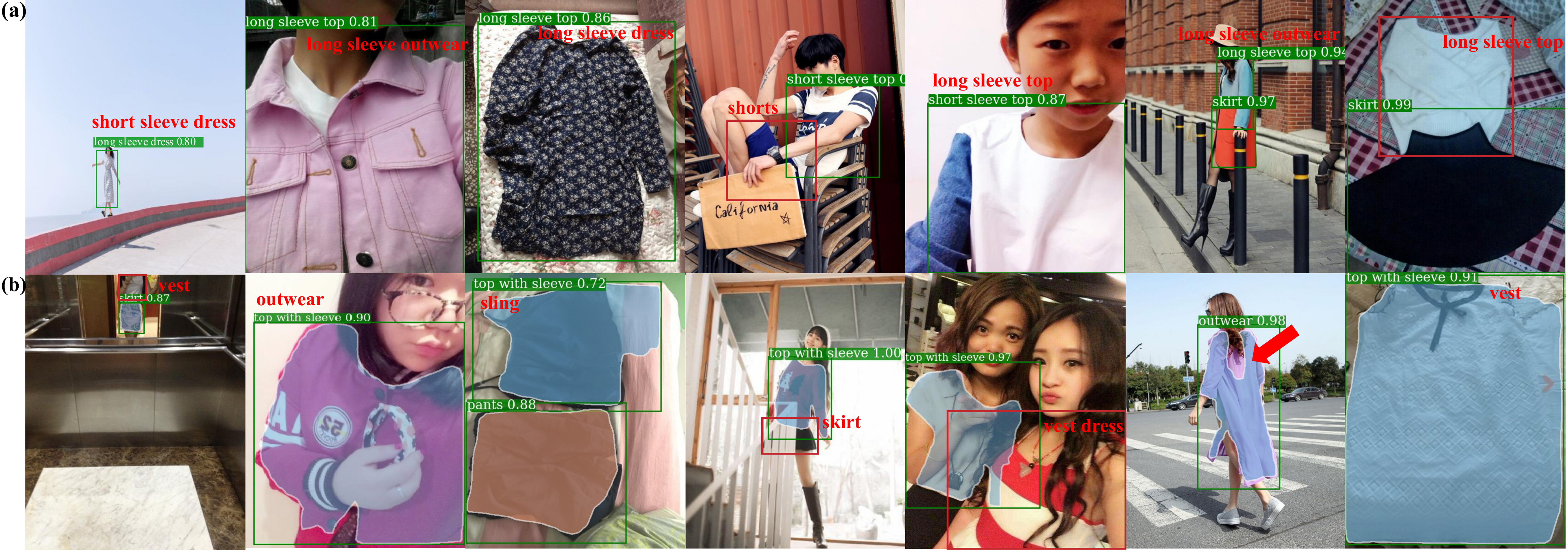}
	\end{center}
	\caption{(a) shows failure cases in clothes detection while (b) shows failure cases in clothes segmentation. In (a) and (b), the missing bounding boxes are drawn in red while the correct category labels are also in red. Inaccurate masks are also highlighted by arrows in (b). For example, clothes fail to be detected or segmented in too small scale, too large scale, large non-rigid deformation, heavy occlusion, large zoom-in, side or back viewpoint.
	}
	\label{fig:failure}
\end{figure*}

In the third stage, MN contains a feature extractor and a similarity learning network for clothes retrieval.
The learned RoI features after the FN component are highly discriminative with respect to clothes category, pose, and mask.
They are fed into MN to obtain features vectors for retrieval,
where $v_1$ and $v_2$ are passed into the similarity learning network to obtain the similarity score between the detected clothing items in $I_1$ and $I_2$.
Specifically, the feature extractor has 4 `conv' layers, one pooling layer, and one fc layer.
%
%Inspired by recent work~\cite{C:PersonSearch}, 
The similarity learning network consists of subtraction and square operator and a fc layer, which estimates the probability of whether two clothing items match or not.
\textbf{Loss Functions.} The parameters $\Theta$ of the Match R-CNN are optimized by minimizing five loss functions, which are formulated as $\min_\Theta\mathcal{L}=\lambda_1\mathcal{L}_{cls}+\lambda_2\mathcal{L}_{box}+\lambda_3\mathcal{L}_{pose}+\lambda_4\mathcal{L}_{mask}+\lambda_5\mathcal{L}_{pair}$, including a cross-entropy (CE) loss $\mathcal{L}_{cls}$ for clothes classification, a smooth loss \cite{C:fast} $\mathcal{L}_{box}$ for bounding box regression, a CE loss $\mathcal{L}_{pose}$ for landmark estimation, a CE loss $\mathcal{L}_{mask}$ for clothes segmentation, and a CE loss $\mathcal{L}_{pair}$ for clothes retrieval. 
Specifically, $\mathcal{L}_{cls}$, $\mathcal{L}_{box}$, $\mathcal{L}_{pose}$, and $\mathcal{L}_{mask}$ are identical as defined in \cite{C:MaskRCNN}. We have $\mathcal{L}_{pair}= -\frac{1}{n}\sum_{i=1}^{n}[y_{i}log(\hat{y}_{i})+(1-y_{i})log(1-\hat{y}_{i})]$,
where $y_{i}=1$ indicates the two items of a pair are matched, otherwise $y_{i}=0$.
% indicates they are not matched.
% clothing item pairs.
%
%$\hat{y_{i}}$ and $1-\hat{y_{i}}$ are the outputs of MN.
\begin{table*}[t]
	\small
	\scalebox{0.9}{
		\begin{tabular}{l|ccc|ccc|ccc|ccc|c}
			\hline
			&\multicolumn{3}{c|}{scale}&\multicolumn{3}{c|}{occlusion}&\multicolumn{3}{c|}{zoom-in}&\multicolumn{3}{c|}{viewpoint}&\multicolumn{1}{c}{overall} \\
			\quad &small&moderate&large&slight&medium&heavy&no&medium&large&no wear&frontal&side or back& \\
			\hline
			\multirow{2}*{$\mathrm{AP_{pt}}$}&0.587&\textbf{0.687}&0.599&\textbf{0.669}&0.631&0.398&\textbf{0.688}&0.559&0.375&0.527&\textbf{0.677}&0.536&0.641 \\
			&0.497&\textbf{0.607}&0.555&\textbf{0.643}&0.530&0.248&\textbf{0.616}&0.489&0.319&0.510&\textbf{0.596}&0.456&0.563\\
			\hline
			\multirow{2}*{$\mathrm{AP^{OKS=0.50}_{pt}}$}&0.780&\textbf{0.854}&0.782&\textbf{0.851}&0.813&0.534&\textbf{0.855}&0.757&0.571&0.724&\textbf{0.846}&0.748&0.820\\
			&0.764&\textbf{0.839}&0.774&\textbf{0.847}&0.799&0.479&\textbf{0.848}&0.744&0.549&0.716&\textbf{0.832}&0.727&0.805\\
			\hline
			\multirow{2}*{$\mathrm{AP^{OKS=0.75}_{pt}}$}&0.671&\textbf{0.779}&0.678&\textbf{0.760}&0.718&0.440&\textbf{0.786}&0.633&0.390&0.571&\textbf{0.771}&0.610&0.728\\
			&0.551&\textbf{0.703}&0.625&\textbf{0.739}&0.600&0.236&\textbf{0.714}&0.537&0.307&0.550&\textbf{0.684}&0.506&0.641\\
			\hline
	\end{tabular}}
	\vspace{3pt}
	\caption{\textbf{Landmark estimation} of Mask R-CNN \cite{C:MaskRCNN} on different validation subsets, including scale, occlusion, zoom-in, and viewpoint. Results of evaluation on visible landmarks only and evaluation on both visible and occlusion landmarks are separately shown in each row. The evaluation metrics are $\mathrm{AP}_{\mathrm{pt}}$, $\mathrm{AP^{OKS=0.50}_{pt}}$, and $\mathrm{AP^{OKS=0.75}_{pt}}$. The best performance of each subset is bold. }
	\label{tab:landmark}
\end{table*}

\textbf{Implementations.}
In our experiments, each training image is resized to its shorter edge of 800 pixels with its longer edge that is no more than 1333 pixels.
Each minibatch has two images in a GPU and 8 GPUs are used for training.
%We use training schedules the same way as Mask R-CNN.
%
For minibatch size 16, the learning rate (LR) schedule starts at $0.02$ and is decreased by a factor of $0.1$ after 8 epochs and then 11 epochs, and finally terminates at 12 epochs. This scheduler is denoted as 1x.
Mask R-CNN adopts 2x schedule for clothes detection and segmentation where `2x' is twice as long as 1x with the LR scaled proportionally. Then It adopts s1x for landmark and pose estimation where s1x scales the 1x schedule by roughly 1.44x.
Match R-CNN uses 1x schedule for consumer-to-shop clothes retrieval.
The above models are trained by using SGD with a weight decay of $10^{-5}$ and momentum of $0.9$.
\par In our experiments, the RPN produces anchors with 3
aspect rations on each level of the FPN pyramid.
 In clothes
detection stream, an RoI is considered positive if its IoU
with a ground truth box is larger than 0.5 and negative otherwise. 
In clothes segmentation stream, positive RoIs with
foreground label are chosen 
while in landmark estimation
stream, positive RoIs with visible landmarks are selected.
We define ground truth box of interest as clothing items
whose style number is $>$ 0 and can constitute matching
pairs. 
In clothes retrieval stream,
RoIs
are selected if their IoU with a ground truth box of interest is
larger than 0.7. 
If RoI features are extracted from landmark
estimation stream, RoIs with visible landmarks are also
selected. 
\par\textbf{Inference.} At testing time, images are resized in the same way as the training stage.
The top 1000 proposals with detection probabilities are chosen for bounding box classification and regression.
Then non-maximum suppression is applied to these proposals.
The filtered proposals are fed into the landmark branch and the mask branch separately.
% for landmark estimation and segmentation.
%
For the retrieval task, each unique detected clothing item in consumer-taken image with highest confidence is selected as query.

\begin{figure}[t]
	\centering
	\includegraphics[width=1\linewidth,height=13.5cm]{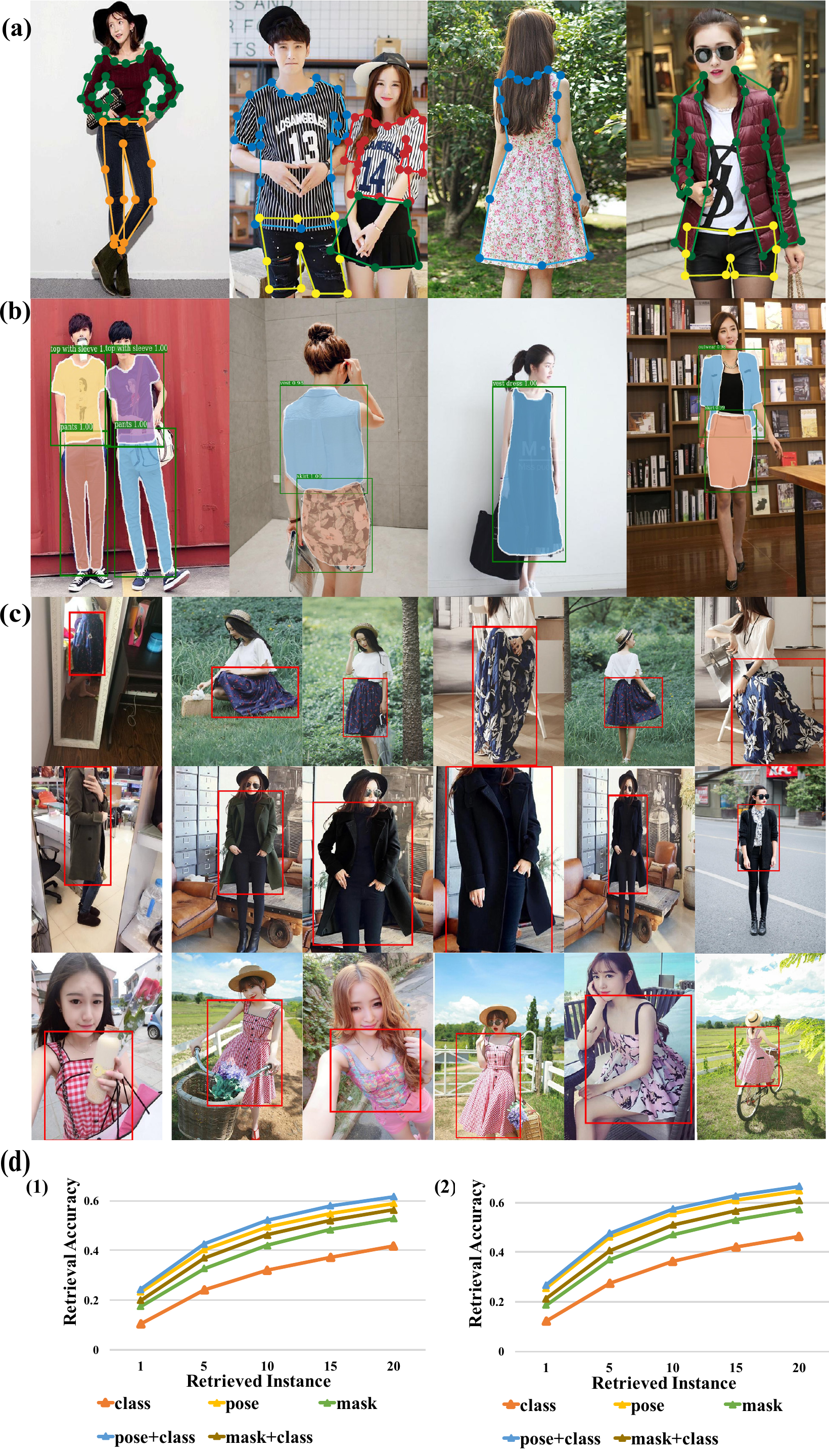}
	\caption{(a) shows results of landmark and pose estimation. (b) shows results of clothes segmentation. (c) shows queries with top-5 retrieved clothing items. The first column is the image from the customer with bounding box predicted by detection module, and the second to the sixth columns show the retrieval results from the store. (d) is the retrieval accuracy of overall query validation set with (1) detected box (2) ground truth box. Evaluation metrics are top-1, -5, -10, -15, and -20 retrieval accuracy.}
	\label{fig:results}
\end{figure}

\section{Experiments}

We demonstrate the effectiveness of DeepFashion2 by evaluating Mask R-CNN \cite{C:MaskRCNN} and Match R-CNN in multiple tasks including clothes detection and classification, landmark estimation, instance segmentation, and consumer-to-shop clothes retrieval.
To further show the large variations of DeepFashion2, the validation set is divided into three subsets according to their difficulty levels in scale, occlusion, zoom-in, and viewpoint.
The settings of Mask R-CNN and Match R-CNN follow Sec.\ref{sec:implementation}. All models are trained in the training set and evaluated in the validation set.

The following sections from \ref{sec:detection} to \ref{sec:retrieval} report results for different tasks, showing that DeepFashion2 imposes significant challenges to both Mask R-CNN and Match R-CNN, which are the recent state-of-the-art systems for visual perception.

\begin{table*}[t]
	\small
	\scalebox{0.9}{
		\begin{tabular}{l|ccc|ccc|ccc|ccc|c}
			\hline
			&\multicolumn{3}{c|}{scale}&\multicolumn{3}{c|}{occlusion}&\multicolumn{3}{c|}{zoom-in}&\multicolumn{3}{c|}{viewpoint}&\multicolumn{1}{c}{overall}\\
			\quad &small&moderate&large&slight&medium&heavy&no&medium&large&no wear&frontal&side or back& \\
			\hline
			$\mathrm{AP_{mask}}$    &0.634&\textbf{0.700}&0.669&\textbf{0.720}&0.674&0.389&\textbf{0.703}&0.627&0.526&0.695&\textbf{0.697}&0.617&0.680\\
			$\mathrm{AP^{IoU=0.50}_{mask}}$ &0.831&\textbf{0.900}&0.844&\textbf{0.900}&0.878&0.559&\textbf{0.899}&0.815&0.663&0.829&\textbf{0.886}&0.843&0.873\\
			$\mathrm{AP^{IoU=0.75}_{mask}}$ &0.765&\textbf{0.838}&0.786&\textbf{0.850}&0.813&0.463&\textbf{0.842}&0.740&0.613&0.792&\textbf{0.834}&0.732&0.812\\
			\hline
	\end{tabular}}
	\vspace{3pt}
	\caption{\textbf{Clothes segmentation} of Mask R-CNN \cite{C:MaskRCNN} on different validation subsets, including scale, occlusion, zoom-in, and viewpoint. The evaluation metrics are $\mathrm{AP}_{\mathrm{mask}}$, $\mathrm{AP^{IoU=0.50}_{mask}}$, and $\mathrm{AP^{IoU=0.75}_{mask}}$. The best performance of each subset is bold.}
	
	\label{tab:parsing}
\end{table*}

\begin{table*}[t]
	\small
	\centering
	\scalebox{0.8}{
		\begin{tabular}{l|ccc|ccc|ccc|ccc|ccc}
			\hline
			&\multicolumn{3}{c|}{scale}&\multicolumn{3}{c|}{occlusion}&\multicolumn{3}{c|}{zoom-in}&\multicolumn{3}{c|}{viewpoint}&\multicolumn{3}{c}{overall}\\
			\quad &small&moderate&large&slight&medium&heavy&no&medium&large&no wear&frontal&side or back&top-1&top-10&top-20\\
			\hline
			\multirow{2}*{class}&0.513&\textbf{0.619}&0.547&\textbf{0.580}&0.556&0.503&\textbf{0.608}&0.557&0.441&0.555&\textbf{0.580}&0.533&0.122&0.363&0.464\\
&0.445&\textbf{0.558}&0.515&\textbf{0.542}&0.514&0.361&\textbf{0.557}&0.514&0.409&0.508&\textbf{0.529}&0.519&0.104&0.321&0.417\\
\hline
\multirow{2}*{pose}&0.695&\textbf{0.775}&0.729&\textbf{0.752}&0.729&0.698&\textbf{0.769}&0.742&0.618&0.725&\textbf{0.755}&0.705&0.255&0.555&0.647\\
&0.619&\textbf{0.695}&0.688&\textbf{0.704}&0.668&0.559&\textbf{0.700}&0.693&0.572&0.682&\textbf{0.690}&0.654&0.234&0.495&0.589\\
\hline
\multirow{2}*{mask}&0.641&\textbf{0.705}&0.663&\textbf{0.688}&0.656&0.645&\textbf{0.708}&0.670&0.556&0.650&\textbf{0.690}&0.653&0.187&0.471&0.573\\
&0.584&\textbf{0.656}&0.632&\textbf{0.657}&0.619&0.512&\textbf{0.663}&0.630&0.541&0.628&\textbf{0.645}&0.602&0.175&0.421&0.529\\
\hline
\multirow{2}*{pose+class}&0.752&\textbf{0.786}&0.733&\textbf{0.754}&0.750&0.728&\textbf{0.789}&0.750&0.620&0.726&\textbf{0.771}&0.719&0.268&0.574&0.665\\
&0.691&\textbf{0.730}&0.705&\textbf{0.725}&0.706&0.605&\textbf{0.746}&0.709&0.582&0.699&\textbf{0.723}&0.684&0.244&0.522&0.617\\
\hline
\multirow{2}*{mask+class} &0.679&\textbf{0.738}&0.685&\textbf{0.711}&0.695&0.651&\textbf{0.742}&0.699&0.569&0.677&\textbf{0.719}&0.678&0.214&0.510&0.607\\
&0.623&\textbf{0.696}&0.661&\textbf{0.685}&0.659&0.568&\textbf{0.708}&0.667&0.566&0.659&\textbf{0.676}&0.657&0.200&0.463&0.564\\
			\hline
	\end{tabular}}
	\vspace{3pt}
	\caption{\textbf{Consumer-to-Shop Clothes Retrieval} of Match R-CNN on different subsets of some validation consumer-taken images.  Each query
item in these images has over 5 identical clothing items in validation commercial images. Results of evaluation on ground truth box and detected box are separately shown in each row. The evaluation metrics are top-20 accuracy. The best performance of each subset is bold.}
	\label{tab:retrieval}
\end{table*}

\subsection{Clothes Detection}\label{sec:detection}
Table~\ref{tab:detection} summarizes the results of clothes detection on different difficulty subsets. We see
% and overall validation set.
%
that the clothes of moderate scale, slight occlusion, no zoom-in, and frontal viewpoint have the highest detection rates.
There are several observations.
First, detecting clothes with small or large scale reduces detection rates. Some failure cases are provided in Fig.\ref{fig:failure}(a) where the item could occupy less than 2\% of the image while some occupies more than 90\% of the image.
Second, in Table~\ref{tab:detection}, it is intuitively to see that heavy occlusion and large zoom-in degenerate performance.
In these two cases, large portions of the clothes are invisible as shown in Fig.\ref{fig:failure}(a).
Third, it is seen in Table~\ref{tab:detection} that the clothing items not on human body  also drop performance. This is because they possess large non-rigid deformations as visualized in the failure cases of Fig.\ref{fig:failure}(a). These variations are not presented in previous object detection benchmarks such as COCO.
Fourth, clothes with side or back viewpoint,  are much more difficult to detect as shown in Fig.\ref{fig:failure}(a).

\subsection{Landmark and Pose Estimation}
Table~\ref{tab:landmark} summarizes the results of landmark estimation.
% on different subsets and overall validation set.
%
The evaluation of each subset is performed in two settings, including visible landmark only (the occluded landmarks are not evaluated), as well as both visible and occluded landmarks.
As estimating the occluded landmarks is more difficult than visible landmarks, the second setting generally provides worse results than the first setting.

In general, we see that Mask R-CNN obtains an overall AP of just 0.563, showing that clothes landmark estimation could be even more challenging than human pose estimation in COCO.
In particular, Table \ref{tab:landmark} exhibits similar trends as those from clothes detection.
For example, the clothing items with moderate scale, slight occlusion, no zoom-in, and frontal viewpoint have better results than the others subsets.
Moreover, heavy occlusion and zoom-in decreases performance a lot.
Some results are given in Fig.\ref{fig:results}(a).

\subsection{Clothes Segmentation}
Table~\ref{tab:parsing} summarizes the results of segmentation.
The performance declines when segmenting clothing items with small and large scale, heavy occlusion, large zoom-in, side or back viewpoint, which is consistent with those trends in the previous tasks.
Some results are given in Fig.\ref{fig:results}(b). 
% clothes detection and landmark estimation tasks. 
Some failure cases are visualized in Fig.\ref{fig:failure}(b).

\subsection{Consumer-to-Shop Clothes Retrieval}\label{sec:retrieval}
Table~\ref{tab:retrieval} summarizes the results of clothes retrieval. The retrieval accuracy is reported in Fig.~\ref{fig:results}(d), where top-1, -5, -10, and -20 retrieval accuracy are shown.
We evaluate two settings in (c.1) and (c.2), when the bounding boxes are predicted by the detection module in Match R-CNN and are provided as ground truths.
% box items are separately evaluated.
%
Match R-CNN achieves a top-20 accuracy of less than 0.7 with ground-truth bounding boxes provided, indicating that the retrieval benchmark is challenging.
% retrieval benchmark.
%
Furthermore, retrieval accuracy drops when using detected boxes, meaning that this is a more realistic setting.
% adds to the difficulty of retrieval task.
%

In Table~\ref{tab:retrieval}, different combinations of the learned features are also evaluated.
In general, the combination of features increases the accuracy. In particular, the learned features from pose and class achieve better results than the other features.
%
%Furthermore,
%
When comparing learned features from pose and mask, we find that the former achieves better results, indicating that landmark locations can be more robust across scenarios.

As shown in Table \ref{tab:retrieval}, the performance declines when small scale, heavily occluded clothing items are presented.
Clothes with large zoom-in achieved the lowest accuracy because only part of clothes are displayed in the image and crucial distinguishable features may be missing.
Compared with clothes on people from frontal view, clothes from side or back viewpoint perform worse due to lack of discriminative features like patterns on the front of tops.
Example queries with top-5 retrieved clothing items are shown in Fig.\ref{fig:results}(c).

\section{Conclusions}
This work represented DeepFashion2, a large-scale fashion image benchmark with comprehensive tasks and  annotations.
DeepFashion2 contains 491K images, each of which is richly labeled with style, scale, occlusion, zooming, viewpoint, bounding box, dense landmarks and pose, pixel-level masks, and pair of images of identical item from consumer and commercial store.
We establish benchmarks covering multiple tasks in fashion understanding, including clothes detection, landmark and pose estimation, clothes segmentation, consumer-to-shop verification and retrieval.
A novel Match R-CNN framework that builds upon Mask R-CNN is proposed to solve the above tasks in end-to-end manner.
Extensive evaluations are conducted in DeepFashion2.

The rich data and labels of DeepFashion2 will definitely facilitate the developments of algorithms to understand fashion images in future work.
We will focus on three aspects.
First, more challenging tasks will be explored with DeepFashion2, such as synthesizing clothing images by using GANs.
Second, it is also interesting to explore multi-domain learning for clothing images, because fashion trends of clothes may change frequently, making variations of clothing images changed.
Third, we will introduce more evaluation metrics into DeepFashion2, such as size, runtime, and memory consumptions of deep models, towards understanding fashion images in real-world scenario.
%------------------------

{\small
\bibliographystyle{ieee}
\bibliography{egbib}
}

\end{document}